\documentclass[10pt,conference]{IEEEtran} 
\IEEEoverridecommandlockouts
\usepackage{cite}
\usepackage{amsmath,amssymb,amsfonts}
\usepackage{algorithmic}
\usepackage{graphicx}
\usepackage{textcomp}
\usepackage{xcolor}

\usepackage{multirow}
\usepackage{subfigure}
\usepackage{graphicx}
\usepackage{enumitem}
\usepackage{graphicx}
\usepackage{balance}
\usepackage[ruled,linesnumbered]{algorithm2e}
\usepackage{float}
\usepackage{booktabs}
\usepackage{flushend}
\usepackage{color}
\usepackage{soul}
\usepackage{url}
\usepackage{makecell}

\def\BibTeX{{\rm B\kern-.05em{\sc i\kern-.025em b}\kern-.08em
    T\kern-.1667em\lower.7ex\hbox{E}\kern-.125emX}}
\begin{document}

% \title{Engineering Secure Self-Adaptive Systems\\ with Bayesian Games}
% \title{Modeling Component-level Secure Self-Adaptive Systems via Bayesian Games}
% \title{Modeling Component-level Secure Self-Adaptation Using Bayesian Games}
% \title{Modeling Component-level Secure Self-Adaptive Systems as Bayesian Games}
% \title{Bayesian Game Based Self-Adaptations for Security}
\title{System Component-Level Self-Adaptations \\ for Security via Bayesian Games}
% Bayesian game + security + self-adaptation

\author{\IEEEauthorblockN{Mingyue Zhang}
\IEEEauthorblockA{\textit{Key Lab of High Confidence Software Technologies (MoE), Peking University},
Beijing, China \\
mingyuezhang@pku.edu.cn}
}

\maketitle

\begin{abstract}
Security attacks present unique challenges to self-adaptive system design due to the adversarial nature of the environment.
However, modeling the system as a single player, as done in prior works in security domain, is insufficient for the system under partial compromise and for the design of fine-grained defensive strategies where the rest of the system with autonomy can cooperate to mitigate the impact of attacks.
To deal with such issues, we propose a new self-adaptive framework incorporating Bayesian game and model the defender (i.e., the system) at the granularity of components in system architecture.
The system architecture model is translated into a \emph{Bayesian multi-player game}, where each component is modeled as an independent player while security attacks are encoded as variant types for the components. 
The defensive strategy for the system is dynamically computed by solving the pure equilibrium to achieve the best possible system utility, improving the resiliency of the system against security attacks. 
%We illustrate our approach using a security web scenario with load balancing and a case study on an inter-domain routing application.

\end{abstract}
\begin{IEEEkeywords}
Self-adaptation; Bayesian game; Security
\end{IEEEkeywords}

\section{Introduction}
A self-adaptive system is designed to be capable of modifying its structure and behavior at run time in response to changes in its  environment and the system itself~\cite{DBLP:conf/dagstuhl/ChengLGIMABBBCSDFGGGKKKLMMMPSTTWW09,DBLP:conf/dagstuhl/LemosGMSALSTVVWBBBBCDDEGGGGIKKLMMMMMNPPSSSSTWW10}.
Achieving \emph{security} in presence of uncertainty is particularly challenging due to the adversarial nature of the environment~\cite{DBLP:conf/icse/ElkhodaryW07,DBLP:conf/icse/DevanbuS00}. 
%: (1) to avoid detection, a typical attacker may attempt to remain hidden while carrying out its actions, and so accurately estimating its objectives and capabilities can be difficult, and (2) the attacker actively attempts to cause as much harm as possible to the system, and so a typical ``average case'' analysis may not be appropriate for making optimal defensive  decisions~\cite{DBLP:conf/memocode/KinneerWFGG19}.
Various game-theory approaches have been explored in the security domain for modeling interactions between the system and attackers as a \emph{game} between a group of \emph{players} (i.e., system and multiple attackers, each as one player) and computing Nash equilibrium strategies for the system to minimize the impact of possible attacks~\cite{DBLP:books/daglib/0040483,DBLP:journals/csur/DoTHKKBRPI17,DBLP:conf/gamesec/FarhangG16,DBLP:conf/memocode/KinneerWFGG19}. 
These methods can be used to (1) model adversarial behaviors 
by malicious attackers~\cite{DBLP:conf/gamesec/FarhangG16,moothedath2020game}, and (2) design reliable defense for the system by using underlying incentive mechanisms to balance perceived risks  
in a mathematically grounded manner~\cite{DBLP:journals/csur/DoTHKKBRPI17,DBLP:journals/csur/PawlickCZ19}. 
Prior works in security 
relying on game theory approaches~\cite{DBLP:books/daglib/0040483,DBLP:journals/csur/DoTHKKBRPI17,DBLP:conf/gamesec/FarhangG16,DBLP:conf/memocode/KinneerWFGG19} have treated the system as an independent player (i.e., defender). Abstracting the entire system (i.e., monolithic modeling) applies to the design of defense strategies at the system level.
However, a potential attacker or several attackers usually attack the system by exploiting the vulnerabilities spread over different parts of the system.
Such monolithic modeling is insufficient for capturing the situations where only a part of the system is compromised while other parts of the system, with their autonomy and capability, can mitigate the impact of the on-going attacks and compensate for security losses. 

In this work, we argue that compared to a coarse one-player abstraction of the complex system, 
modeling the defender under security attacks at the granularity of \emph{components} is more expressive compared to monolithic modeling, in that it allows the design of fine-grained defensive strategies for the system under partial compromise.
Our approach to the component modeling approach is a trade-off between the level of details and level of abstraction to appropriately portray aforementioned attack situations. Furthermore, we advocate focusing on the system modeling by encoding the on-going attacks on the component as component behavior deviations, as an alternative way instead of modeling attackers themselves as separate players. 

To this end, we pioneer a new self-adaptive framework that leverages \emph{Bayesian games} at the granularity of \emph{components} at the system architecture level. 
Specifically, each essential component will be separately modeled as a player.  
Under attacks, one or more components with vulnerabilities might be exploited with probability by the attackers to deliberately perform harmful actions (i.e., turning into a malicious type). The various security attacks these components might be subject to are encoded as different \emph{types} for players, the way of expressing uncertainty from the Bayesian approach. The rest of the components could form a coalition to fight against those potentially uncooperative 
components. 
The architecture model of the system and the security attacks on components are translated into a Bayesian game structure. 
Then, the adaptive defensive strategy for the system is dynamically computed by solving a pure equilibrium, to achieve the best possible system utility under all assignments of the components to their possible types (i.e., in the presence of security attacks).

% \section{
% A Novel Self-Adaptive Framework
% }
\section{Bayesian Game Extended MAPE-K Loop}
\begin{figure}[!htbp]
\setlength{\abovecaptionskip}{-0.1cm}  
\setlength{\belowcaptionskip}{-1cm}  
  \centering
  \makebox[\columnwidth][c]{
\includegraphics[width=3.3in,height=2.1in]{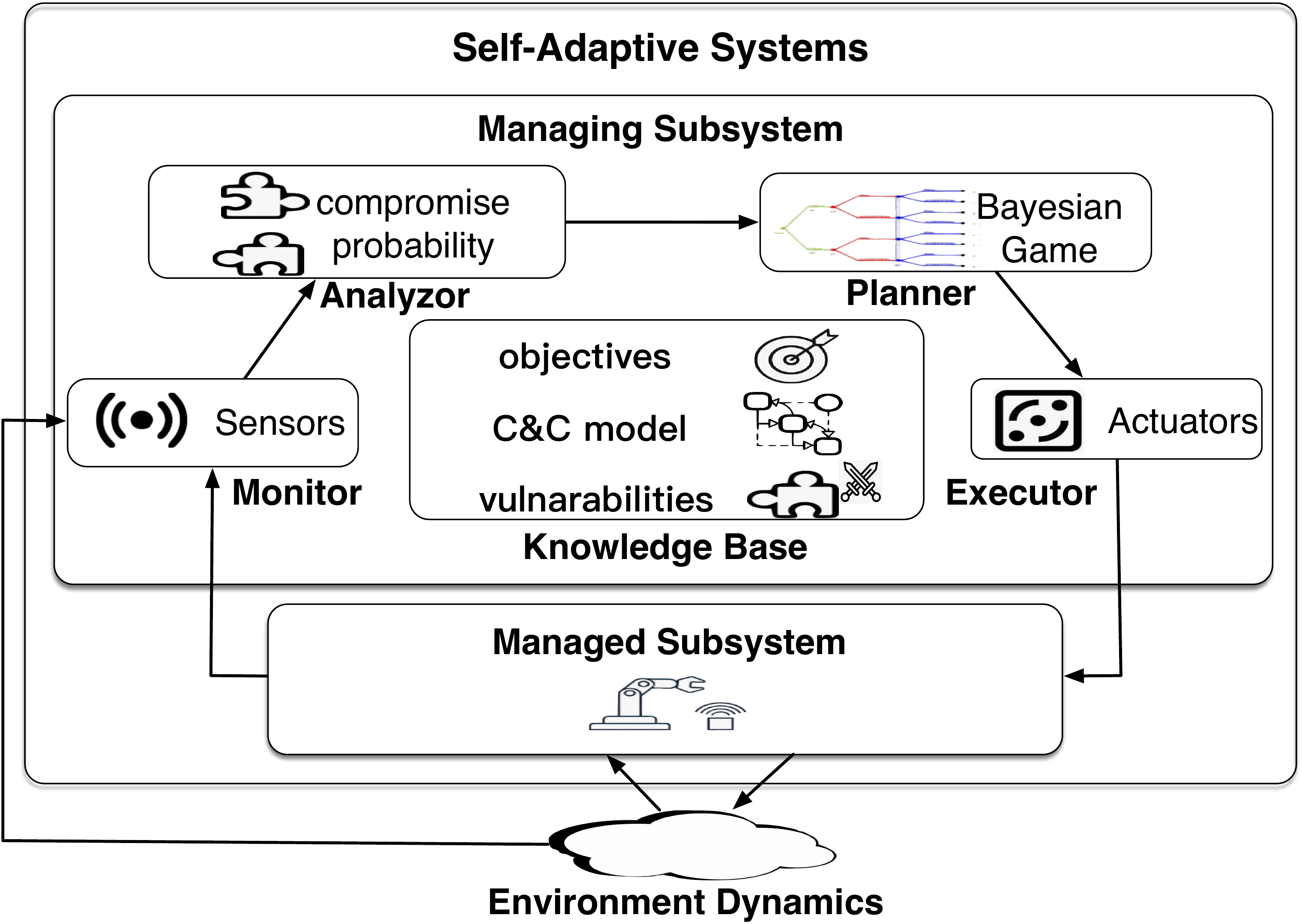}
}
\caption{Self-Adaptive Framework.}% incorporating Bayesian Game Theory.}
\label{overviewfigure}
\end{figure}

% or Game Enabled MAPE-K Loop
% From MAPE-K to Game
\label{overview}
We propose a new self-adaptive framework incorporating Bayesian Game. 
Adaptation behaviors build on the Nash equilibrium from unexpected attacks and
are achieved by elaborating the widely adopted mechanism of the MAPE-K (Monitoring, Analysis, Planning, Execution, Knowledge) loop~\cite{MAPEJeffrey,DBLP:conf/icse/WeynsIS13,braberman2015morph}, shown in Figure~\ref{overviewfigure}. 
Concretely, \emph{Knowledge Base} stores the necessary information for the sake of self-adaptation, including (1) the component and connector model of the managed subsystem and its action space for each component, (2) system objectives usually defined as the quality attributes quantified by the utility, and (3) component vulnerabilities with potential behavior deviations that can be exploited by the potential attacks. \emph{Monitor} gathers and synthesizes the on-going attacks information through sensors and saves information in the Knowledge Base.  \emph{Analyzer} performs analysis and further checks whether certain components are attacked with probabilities; potential deviated malicious actions are identified;
the rewards for the attack are estimated, based on the knowledge about component vulnerabilities and system objectives. \emph{Planner} generates one or a set of adaptation actions by automatically solving the Bayesian Game transformed with the input of potential attacks from the Analyzer and architectural model of the managed subsystem along with the system objectives from the \emph{Knowledge Base}.
Then, adaptations from equilibrium are enacted by \emph{Executor}  on the managed subsystem through actuators.

\vspace{-0.3cm}
\vspace{-0.4cm}

\section{Bayesian Game via Model Transformation}
\label{transformation}
We define the system under attacks, and transform the system architecture with on-going attacks into a component-based Bayesian game. 
Solving the game with equilibrium is to find the adaptation strategy.

\noindent\textbf{Component-based System:} 
A system component is an independent and replaceable part of a system that fulfills a clear function in the context of a well-defined architecture.
Components forming architectural structures will affect different quality attributes. A system is defined as $S = \langle C, A, Q\rangle$, where $C$ is a set of components;  $A$ is a set of joint actions available to component $i$; $Q$ is a set of quality attributes a system is interested in.  
Each component is trying to make the right reaction to maximize the system utility. 
% \cite{javier2016analyzing,javier2013analyzing}.
% essentially like a rational player in the game theory. 
Naturally, a system under normal operation could be viewed as a cooperative game dealing with how coalitions interact \cite{javier2016analyzing,javier2013analyzing}.

%\vspace{0.15cm}
\noindent\textbf{Modeling Utility as Payoff:} 
The payoff among those players is allocated by the utility from quality attributes. 
It is straightforward for developers to design a system-level utility. 
However, due to the different roles of the components and the complex relationship between them, it is complicated and sometimes untraceable to manually design an appropriate component-level payoff function.
The \emph{Shapley value}, a solution of fairly distributing both gains and costs to several players working in coalition proportional to their marginal contributions \cite{shapley1953vaule,Osborne1994A,levinger2020computing}, is used to automatically decompose the system-level utility into the component-level payoff.

\noindent\textbf{Component-based Attacks:} 
Instead of modeling an attacker or several attackers with possible complex behaviors over different parts of the system,
we model the on-going attacks $ATT$ the system is enduring at the component level since the vulnerabilities of the components as well as their potential behavior deviations are comparatively easy to observe and be analyzed.
The security attacks on the system is formally defined as a tuple $ATT = \langle C_{att}, A_{att}, P_{att},R_{att} \rangle$, where $C_{att}$ is the set of components affected by the attacks; $A_{att}$ denotes a set of joint actions controlled by attacks on compromised components; $P_{att} = \{p_1,...,p_m\}$ is a set of probability where $p_i$ is the probability of component $i$ being successfully compromised; $R_{att}$ is the reward for attacks.

\noindent\textbf{Translation into a Bayesian game:} With the definition of the system on the component level and the definition of the attacks, a system under security attacks is converted into a Bayesian game $B = \langle P, A, \Theta, U, \rho \rangle$, where $P$ is a set of players; $A$ is a set of actions; $\Theta$ is a set of types for each player $i: \theta_i \in \Theta_i$; $U$ is a payoff function for each player determined by the types of all players and actions they choose; $\rho$ is a probability distribution $\rho(\theta_1,...,\theta_n)$ over types.

The game translation follows five steps: 
1) each component $c\in C$ is separately modeled as an independent player; 
2) components potentially affected by attacks $C_{att}\subseteq C$ will be associated with two types (i.e, \emph{normal} and \emph{malicious}) while the remaining components $C - C_{att}$ are \emph{normal} type;
3) the probability distribution for a player $i$ over two types is $\rho(p_i,1-p_i)$ as defined in $P_{att}$; 
4) the action space of player $i$ under security attacks is $A_i \cup A_{att}$; 
5) the payoff for players in normal type will be allocated with system utility by the \emph{shapley value method}, while components in malicious type performing harmful actions will be assigned with utility the on-going attacks obtain by achieving their own goals. 
The game constructed is put into a game solver (i.e., Gambit \cite{gambit}), to find Nash equilibria, which, in essence, is the best reaction as the adaptation response for the system to potential attacks.

%Note that this definition can be easily extended for the situation where a component is simultaneously compromised by different attackers with multiple types.  
%Besides, the game solver we adopted in this work is \emph{Gambit} \cite{gambit}, a collection of tools for building game models, computing game equilibrium and analyzing game results, to efficiently model the Bayesian game translated by the above steps and automatically figure out the equilibrium strategy as the adaptation response.

\section{Conclusion and Future Work}
\label{sec:conclusions}
In this paper, we have proposed a new framework for self-adaptive systems by adopting Bayesian game theory and modeled the system under security attacks as a multi-player game at system architecture level. Its applicability and superiority have been demonstrated in a security web scenario with load balancing and a case study on an inter-domain routing application \footnote{\url{https://github.com/GeorgeDUT/GamePlusAdaptation2ICSEsrc}}.
In future, we are planning to evaluate our framework in a realistic industrial control system with adaptive behaviors~\cite{mathurTippenhauer,mawAdepu2019ics,chen2019learning,adepu2020control} by  constructing the game in an automated way and supporting Architecture Description Interchange Language, such as acme~\cite{DBLP:conf/cascon/GarlanMW97}.

\section*{Acknowledgment}
This work is supported by the National Natural Science Foundation of China under Grant No.61620106007 and No.61751210.

% \section*{References}

\bibliographystyle{IEEEtran}
\bibliography{sample-base}

\begin{thebibliography}{10}
\providecommand{\url}[1]{#1}
\csname url@samestyle\endcsname
\providecommand{\newblock}{\relax}
\providecommand{\bibinfo}[2]{#2}
\providecommand{\BIBentrySTDinterwordspacing}{\spaceskip=0pt\relax}
\providecommand{\BIBentryALTinterwordstretchfactor}{4}
\providecommand{\BIBentryALTinterwordspacing}{\spaceskip=\fontdimen2\font plus
\BIBentryALTinterwordstretchfactor\fontdimen3\font minus
  \fontdimen4\font\relax}
\providecommand{\BIBforeignlanguage}[2]{{%
\expandafter\ifx\csname l@#1\endcsname\relax
\typeout{** WARNING: IEEEtran.bst: No hyphenation pattern has been}%
\typeout{** loaded for the language `#1'. Using the pattern for}%
\typeout{** the default language instead.}%
\else
\language=\csname l@#1\endcsname
\fi
#2}}
\providecommand{\BIBdecl}{\relax}
\BIBdecl

\bibitem{DBLP:conf/dagstuhl/ChengLGIMABBBCSDFGGGKKKLMMMPSTTWW09}
B.~H.~C. Cheng and et~al., ``Software engineering for self-adaptive systems:
  {A} research roadmap,'' in \emph{Software Engineering for Self-Adaptive
  Systems [outcome of a Dagstuhl Seminar]}, 2009, pp. 1--26.

\bibitem{DBLP:conf/dagstuhl/LemosGMSALSTVVWBBBBCDDEGGGGIKKLMMMMMNPPSSSSTWW10}
R.~de~Lemos and et~al., ``Software engineering for self-adaptive systems: {A}
  second research roadmap,'' in \emph{Software Engineering for Self-Adaptive
  Systems {II} - International Seminar, Dagstuhl Castle, Germany, October
  24-29, 2010 Revised Selected and Invited Papers}, 2010, pp. 1--32.

\bibitem{DBLP:conf/icse/ElkhodaryW07}
A.~M. Elkhodary and J.~Whittle, ``A survey of approaches to adaptive
  application security,'' in \emph{2007 {ICSE} Workshop on Software Engineering
  for Adaptive and Self-Managing Systems, {SEAMS} 2007, Minneapolis Minnesota,
  USA, May 20-26, 2007}, 2007, p.~16.

\bibitem{DBLP:conf/icse/DevanbuS00}
P.~T. Devanbu and S.~G. Stubblebine, ``Software engineering for security: a
  roadmap,'' in \emph{22nd International Conference on on Software Engineering,
  Future of Software Engineering Track, {ICSE} 2000, Limerick Ireland, June
  4-11, 2000}, 2000, pp. 227--239.

\bibitem{DBLP:books/daglib/0040483}
M.~Tambe, \emph{Security and Game Theory - Algorithms, Deployed Systems,
  Lessons Learned}.\hskip 1em plus 0.5em minus 0.4em\relax Cambridge University
  Press, 2012.

\bibitem{DBLP:journals/csur/DoTHKKBRPI17}
\BIBentryALTinterwordspacing
C.~T. Do, N.~H. Tran, C.~S. Hong, C.~A. Kamhoua, K.~A. Kwiat, E.~Blasch,
  S.~Ren, N.~Pissinou, and S.~S. Iyengar, ``Game theory for cyber security and
  privacy,'' \emph{{ACM} Comput. Surv.}, vol.~50, no.~2, pp. 30:1--30:37, 2017.
  [Online]. Available: \url{https://doi.org/10.1145/3057268}
\BIBentrySTDinterwordspacing

\bibitem{DBLP:conf/gamesec/FarhangG16}
\BIBentryALTinterwordspacing
S.~Farhang and J.~Grossklags, ``Flipleakage: {A} game-theoretic approach to
  protect against stealthy attackers in the presence of information leakage,''
  in \emph{Decision and Game Theory for Security - 7th International
  Conference, GameSec 2016, New York, NY, USA, November 2-4, 2016,
  Proceedings}, 2016, pp. 195--214. [Online]. Available:
  \url{https://doi.org/10.1007/978-3-319-47413-7\_12}
\BIBentrySTDinterwordspacing

\bibitem{DBLP:conf/memocode/KinneerWFGG19}
C.~Kinneer, R.~Wagner, F.~Fang, C.~{Le Goues}, and D.~Garlan, ``Modeling
  observability in adaptive systems to defend against advanced persistent
  threats,'' in \emph{Proceedings of the 17th {ACM-IEEE} International
  Conference on Formal Methods and Models for System Design, {MEMOCODE} 2019,
  La Jolla, CA, USA, October 9-11, 2019}, 2019, pp. 10:1--10:11.

\bibitem{moothedath2020game}
S.~Moothedath, D.~Sahabandu, J.~Allen, A.~Clark, L.~Bushnell, W.~Lee, and
  R.~Poovendran, ``A game-theoretic approach for dynamic information flow
  tracking to detect multi-stage advanced persistent threats,'' \emph{IEEE
  Transactions on Automatic Control}, 2020.

\bibitem{DBLP:journals/csur/PawlickCZ19}
J.~Pawlick, E.~Colbert, and Q.~Zhu, ``A game-theoretic taxonomy and survey of
  defensive deception for cybersecurity and privacy,'' \emph{{ACM} Comput.
  Surv.}, vol.~52, no.~4, pp. 82:1--82:28, 2019.

\bibitem{MAPEJeffrey}
\BIBentryALTinterwordspacing
J.~O. Kephart and D.~M. Chess, ``The vision of autonomic computing,''
  \emph{{IEEE} Computer}, vol.~36, no.~1, pp. 41--50, 2003. [Online].
  Available: \url{https://doi.org/10.1109/MC.2003.1160055}
\BIBentrySTDinterwordspacing

\bibitem{DBLP:conf/icse/WeynsIS13}
D.~Weyns, M.~U. Iftikhar, and J.~S{\"{o}}derlund, ``Do external feedback loops
  improve the design of self-adaptive systems? a controlled experiment,'' in
  \emph{Proceedings of the 8th International Symposium on Software Engineering
  for Adaptive and Self-Managing Systems, {SEAMS} 2013, San Francisco, CA, USA,
  May 20-21, 2013}, 2013, pp. 3--12.

\bibitem{braberman2015morph}
V.~Braberman, N.~D'Ippolito, J.~Kramer, D.~Sykes, and S.~Uchitel, ``Morph: A
  reference architecture for configuration and behaviour self-adaptation,'' in
  \emph{Proceedings of the 1st International Workshop on Control Theory for
  Software Engineering}.\hskip 1em plus 0.5em minus 0.4em\relax ACM, 2015, pp.
  9--16.

\bibitem{javier2016analyzing}
J.~C{\'{a}}mara, G.~A. Moreno, D.~Garlan, and B.~R. Schmerl, ``Analyzing
  latency-aware self-adaptation using stochastic games and simulations,''
  \emph{{ACM} Trans. Auton. Adapt. Syst.}, vol.~10, no.~4, pp. 23:1--23:28,
  2016.

\bibitem{javier2013analyzing}
J.~C{\'{a}}mara, D.~Garlan, G.~A. Moreno, and B.~R. Schmerl, ``Analyzing
  self-adaptation via model checking of stochastic games,'' in \emph{Software
  Engineering for Self-Adaptive Systems {III.} Assurances - International
  Seminar, Revised Selected and Invited Papers}, ser. Lecture Notes in Computer
  Science, vol. 9640.\hskip 1em plus 0.5em minus 0.4em\relax Springer, 2013,
  pp. 154--187.

\bibitem{shapley1953vaule}
L.~S. Shapley, ``A value for n-person games,'' \emph{In Contributions to the
  Theory of Games}, vol. vol. 2, 1953.

\bibitem{Osborne1994A}
M.~J. Osborne and A.~Rubinstein, ``A course in game theory,'' \emph{MIT Press
  Books}, vol.~1, 1994.

\bibitem{levinger2020computing}
C.~Levinger, N.~Hazon, and A.~Azaria, ``Computing the shapley value for
  ride-sharing and routing games,'' in \emph{Proceedings of the 19th
  International Conference on Autonomous Agents and MultiAgent Systems}, 2020,
  pp. 1895--1897.

\bibitem{gambit}
R.~D. McKelvey, A.~M. McLennan, and T.~L. Turocy, ``Gambit: Software tools for
  game theory, version 16.0.1,'' 2018-02, \url{http://www.gambit-project.org}.

\bibitem{mathurTippenhauer}
A.~P. Mathur and N.~O. Tippenhauer, ``{SWaT}: {A} water treatment testbed for
  research and training on {ICS} security,'' in \emph{2016 International
  Workshop on Cyber-physical Systems for Smart Water Networks (CySWater)},
  April 2016, pp. 31--36.

\bibitem{mawAdepu2019ics}
A.~Maw, S.~Adepu, and A.~Mathur, ``{ICS-BlockOpS:} blockchain for operational
  data security in industrial control system,'' \emph{Pervasive and Mobile
  Computing}, vol.~59, p. 101048, 2019.

\bibitem{chen2019learning}
Y.~Chen, C.~M. Poskitt, J.~Sun, S.~Adepu, and F.~Zhang, ``Learning-guided
  network fuzzing for testing cyber-physical system defences,'' in \emph{2019
  34th IEEE/ACM International Conference on Automated Software Engineering
  (ASE)}.\hskip 1em plus 0.5em minus 0.4em\relax IEEE, 2019, pp. 962--973.

\bibitem{adepu2020control}
S.~Adepu, F.~Brasser, L.~Garcia, M.~Rodler, L.~Davi, A.-R. Sadeghi, and
  S.~Zonouz, ``Control behavior integrity for distributed cyber-physical
  systems,'' in \emph{2020 ACM/IEEE 11th International Conference on
  Cyber-Physical Systems (ICCPS)}.\hskip 1em plus 0.5em minus 0.4em\relax IEEE,
  2020, pp. 30--40.

\bibitem{DBLP:conf/cascon/GarlanMW97}
\BIBentryALTinterwordspacing
D.~Garlan, R.~T. Monroe, and D.~Wile, ``Acme: an architecture description
  interchange language,'' in \emph{Proceedings of the 1997 conference of the
  Centre for Advanced Studies on Collaborative Research, November 10-13, 1997,
  Toronto, Ontario, Canada}, 1997, p.~7. [Online]. Available:
  \url{https://dl.acm.org/citation.cfm?id=782017}
\BIBentrySTDinterwordspacing

\end{thebibliography}

\vspace{12pt}
% \color{red}
% IEEE conference templates contain guidance text for composing and formatting conference papers. Please ensure that all template text is removed from your conference paper prior to submission to the conference. Failure to remove the template text from your paper may result in your paper not being published.

\end{document}